\documentclass[letterpaper]{article} 
\usepackage[preprint]{aaai2027}  
\usepackage[hyphens]{url}  
\usepackage{graphicx} 
\usepackage{natbib}  
\usepackage{caption} 
\usepackage{amssymb} 
\usepackage{amsmath}
\usepackage{booktabs}

\title{Regional Anchor Routing for Multi-Object Motion Transfer}
\author{
    Yuze Li\textsuperscript{\rm 1}\quad
    Fuchen Long\textsuperscript{\rm 2}\quad
    Xinlong Sun\textsuperscript{\rm 2}\quad
    Cheng Yan\textsuperscript{\rm 1$\S$}\quad
    Xinyu Zhang\textsuperscript{\rm 3}\quad
    Xi Chen\textsuperscript{\rm 2}\quad
}
\affiliations{
    \textsuperscript{\rm 1} Tianjin University,
    \textsuperscript{\rm 2} Smart Creation Platform Department, Online Video BU, Tencent,
    \textsuperscript{\rm 3} University of Auckland \\ 
}

\begin{document}

\maketitle

\begingroup
\renewcommand{\thefootnote}{$*$}
\footnotetext{Work done during the Tencent internship.}
\renewcommand{\thefootnote}{$\S$}
\footnotetext{Corresponding Author.}
\endgroup

\begin{figure*}[t]
\centering
\includegraphics[width=\textwidth]{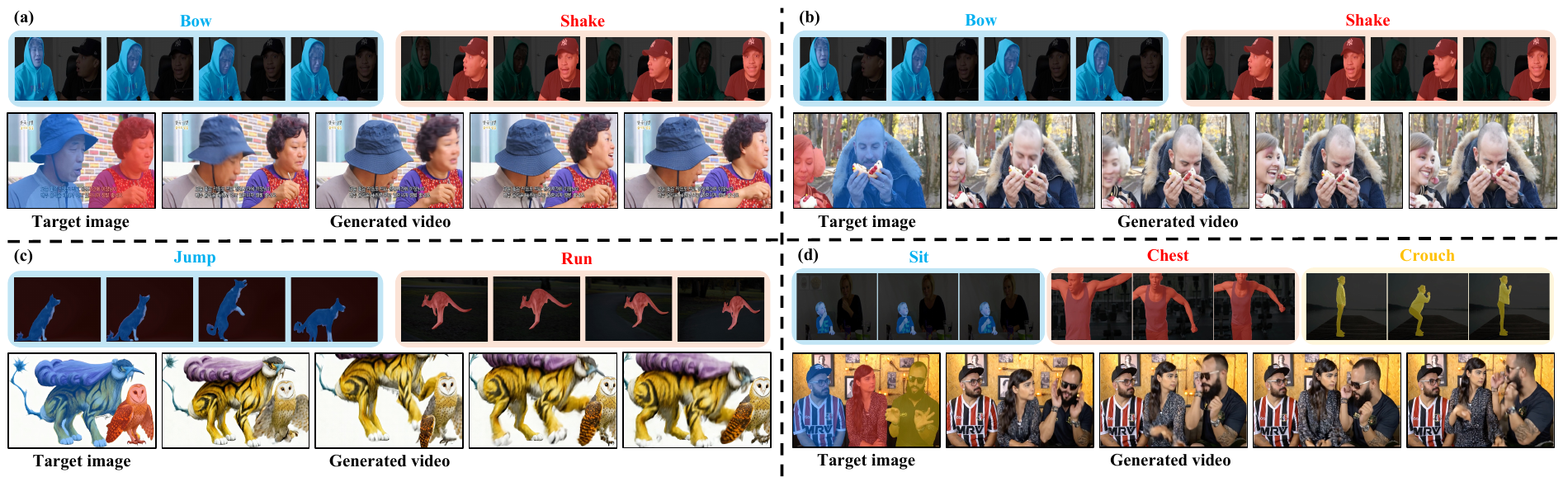}
\caption{Compositional motion transfer with RART. Masked reference videos provide object-wise motions, which are assigned to colored target regions and jointly rendered in the generated video. The examples show (a) and (b) motion reassignment, (c) cross-species transfer, and (d) multi-object composition.}
\label{fig:teaser}
\end{figure*}

\begin{abstract}
Compositional motion transfer requires a model to extract motions from multiple reference videos, and apply each motion to a user-specified object in one target scene.
Existing methods normally struggle to reliably assign different reference motions to the correct target objects, thus a motion intended for one object may affect another. Meanwhile, some motion-specific tuning approaches are still facing another issue: they must run a new optimization whenever the reference video changes, making repeated transfers inefficient.
To address these limitations, we propose RART, a framework that unifies object-wise motion extraction, target-region anchoring, and frame-aligned injection into a frozen video Diffusion Transformer for multi-object motion transfer. 
For each reference motion, RART extracts temporally ordered features from the masked reference video, and then conditions them on the target first-frame mask to form mask-anchored motion tokens. 
Frame-aligned motion cross-attention allows each generated frame to read only the motion tokens at the corresponding time slots, composing all object motions in one denoising process. During training, a motion-attention routing loss encourages each target region to attend to its corresponding motion tokens and reduces interference between objects. 
Therefore, RART can handle novel reference-target combinations without per-video optimization. Experiments show that RART achieves accurate motion preservation, efficient inference, and strong compositional generalization.

\begin{links}
\link{Project Page}{https://ethan-li123.github.io/RART_page/}
\end{links}

\end{abstract}

\section{Introduction}

Diffusion Transformers now generate high-quality videos from text and images \cite{ho2020ddpm,rombach2022ldm,peebles2023dit,blattmann2023svd,kong2024hunyuanvideo,wanteam2025wan}, yet fine-grained object-wise motion control still remains challenging. 
Such a control paradigm is crucial for character animation, video editing \cite{coinve2026,editagent2026}, and multi-subject video creation, where objects usually need to follow different existing motion patterns.

Here, we study the problem of compositional motion transfer: given a target image, multiple reference video-mask pairs, and target object masks, the model transfers each reference motion to its assigned target object in the image while preserving the target image's original appearance and background.
Compared with single-object animation~\cite{karras2023dreampose,xu2023magicanimate,hu2023animateanyone}, this setting introduces three coupled requirements. The model must isolate the desired motion from each masked reference without carrying over its appearance or source layout, assign each extracted motion to the correct target region, and ensure that each object follows its reference motion at the correct time step.
Existing approaches address only part of these requirements. 
Explicit-control methods typically rely on poses, optical flow, depth, trajectories, or camera motion~\cite{wang2023videocomposer,zhang2023controlvideo,wang2023motionctrl,yin2023dragnuwa,ma2023followyourpose,chu2025wanmove,2025motionpro} to control motion. 
The signals provide effective structural guidance when reliable priors are available, but obtaining and aligning them across objects often requires task-specific extraction or annotation. 
Implicit approaches transfer motion directly from reference videos without converting it into explicit pose or optical-flow conditions, offering greater flexibility. 
Among them, training-free methods reuse diffusion features or attention for zero-shot transfer~\cite{yatim2023spacetime,ku2024anyv2v,ling2024motionclone}, but mainly handle a single motion source and lack object-wise motion reassignment. 
Training-based methods learn motion representations, but mostly focus on single-object transfer~\cite{zhao2023motiondirector,jeong2023vmc,ouyang2024i2vedit,zhang2025flexiact} and require retraining for each new motion~\cite{zhang2025flexiact}, limiting their generalization. 
Recent studies have explored feed-forward implicit motion transfer~\cite{resslerantal2025dismo,yan2026scail2}. 
DisMo focuses on single-object transfer, and SCAIL-2 does not support motion transfer from multiple independent reference videos. 

To address these limitations, we propose RART, a newly-minted framework that factorizes compositional motion transfer into object-wise motion extraction, target-region anchoring, and frame-aligned injection. For each reference video, the reference mask isolates the desired motion source, and a pretrained motion encoder extracts a temporally ordered feature sequence. RART projects these features into the DiT token space and conditions them on the corresponding target mask, producing motion tokens that encode both the reference motion and its target-region anchor. 
Motion tokens are then grouped by latent-frame index, and concatenated across objects within each group. 
Tokens associated with different video frames remain separate. 
At each denoising step, the video latent tokens attend to the motion tokens belonging to corresponding frame index, not only preserving alignment between each reference motion and its target object's motion but also allowing multiple reference motions to be injected jointly. 
During training, a motion-attention routing loss further discourages each target region from attending to motion tokens of other objects, while preserving attention to its own tokens. After optimization, RART can transfer motions from new reference videos without video-specific optimization. Representative results are shown in Figure~\ref{fig:teaser}.

Our contributions are summarized as follows:
\begin{itemize}
    \item We propose RART, a framework for compositional multi-object motion transfer on a frozen video Diffusion Transformer, supporting flexible reference-target combinations without per-video optimization.
    \item We introduce mask-anchored motion encoding and frame-aligned motion cross-attention, separating motion content from its target location and preserving temporal correspondence during control injection, respectively.
    \item We propose motion-attention routing loss, which encourages each target region to attend to its assigned motion tokens and suppresses cross-object information leakage.
\end{itemize}

\section{Related Work}

\paragraph{Video Diffusion Models.}
Diffusion models have become the dominant backbone for visual generation \cite{song2019generative,ho2020ddpm,song2021sde,song2021ddim,dhariwal2021diffusion,ho2022classifierfree,rombach2022ldm,peebles2023dit,esser2024sdit,lipman2023flowmatching}. Recent video generation systems extend these priors with temporal modules, latent video diffusion, or transformer-based video modeling \cite{ho2022videodiffusion,singer2022makeavideo,ho2022imagenvideo,villegas2022phenaki,hong2022cogvideo,he2022lvdm,wang2023modelscope,chen2023videocrafter,blattmann2023svd,brooks2024sora,kong2024hunyuanvideo,long2024vs,wanteam2025wan}. These models provide strong priors for realistic video synthesis, but text or image conditions alone are still insufficient for fine-grained object-wise motion control, which should be emphasized in research field.

\paragraph{Explicit Motion Transfer.}
One line of controllable video generation relies on explicit motion conditions such as poses, trajectories, depth, optical flow, or camera motion. VideoComposer \cite{wang2023videocomposer} uses compositional spatial and temporal conditions for motion controllability. 
ControlVideo \cite{zhang2023controlvideo} adapts structural controls to text-to-video generation.
MotionCtrl \cite{wang2023motionctrl} separates camera and object motion control.
DragNUWA \cite{yin2023dragnuwa} integrates text, image, and trajectory conditions for fine-grained control.
Wan-Move \cite{chu2025wanmove} uses latent trajectory guidance for motion-controllable generation. Pose-guided animation methods drive a target subject according to a frame-wise pose sequence~\cite{ma2023followyourpose,karras2023dreampose,xu2023magicanimate,hu2023animateanyone}. 
In summary, these methods are effective when reliable explicit structures are available, but they introduce extra extraction and alignment costs and often specialize to particular object categories. 
RART, in contrast, extracts motion directly from reference videos, and can be applied to cross-object motion transfer.

\begin{figure*}[t]
\centering
\includegraphics[width=1\textwidth]{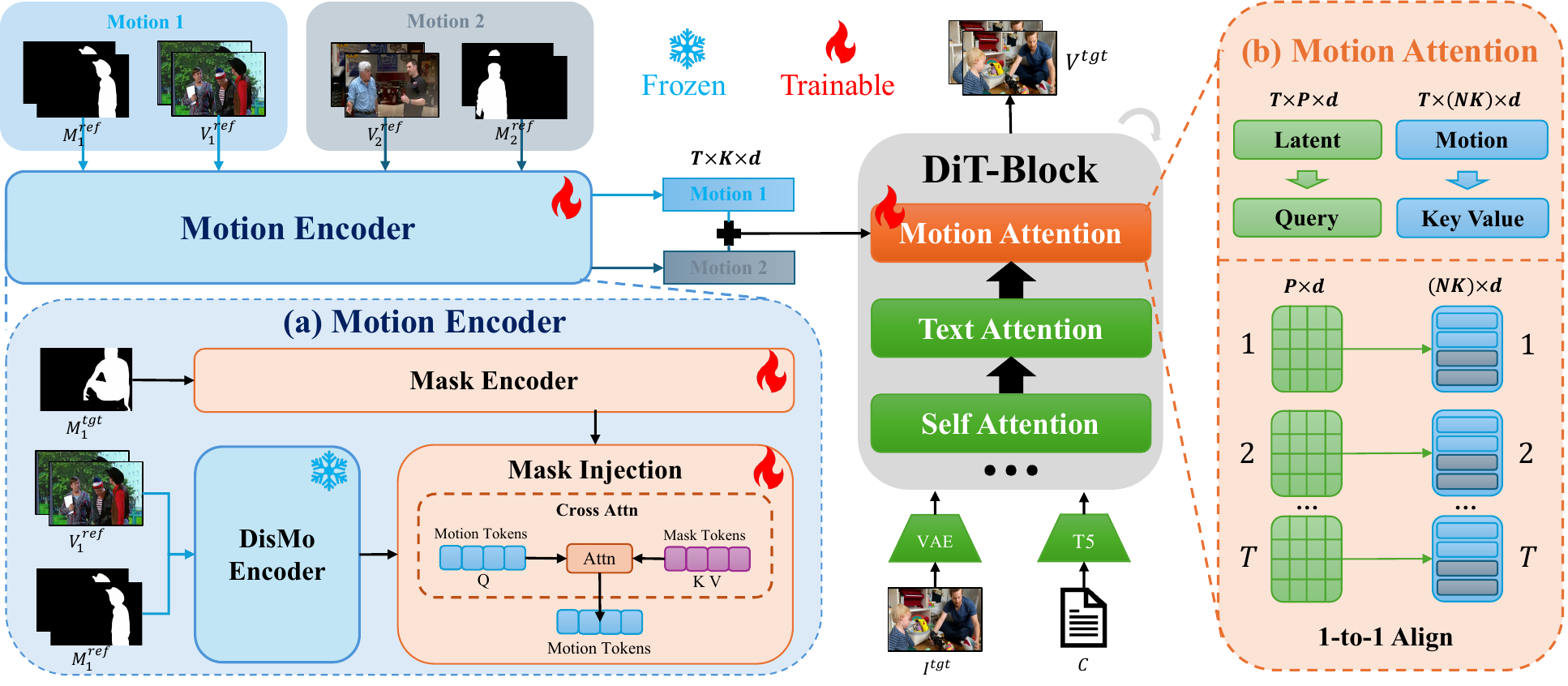}
\caption{Overview of RART. (a) The motion encoder extracts temporally ordered features from each masked reference video and injects the corresponding target-mask anchor to produce object-wise motion tokens. (b) Frame-aligned motion cross-attention injects the motion tokens of each latent frame into the frozen DiT.}
\label{fig:method}
\end{figure*}

\paragraph{Implicit Motion Transfer.}
Another line of work transfers motion through implicit features, which can be divided into two categories.
Training-free approaches keep model parameters fixed but exploit pretrained representations in different ways. 
Space-Time Diffusion Features~\cite{yatim2023spacetime} guides sampling with an internal feature loss.
AnyV2V~\cite{ku2024anyv2v} injects source-video features into an image-to-video process, and MotionClone~\cite{ling2024motionclone} reuses sparse temporal-attention weights. 
These methods avoid reference-specific training but generally process one motion source at a time. 
In comparison, training-based methods~\cite{zhao2023motiondirector,ouyang2024i2vedit} fit motion representations to individual inputs. For instance, MotionDirector~\cite{zhao2023motiondirector} learns spatial and temporal LoRAs. VMC~\cite{jeong2023vmc} adapts temporal layers through motion distillation. 
FlexiAct optimizes frequency-aware embeddings for each reference video~\cite{zhang2025flexiact}. FlexiMMT further extends FlexiAct to multiple objects by learning object-wise motion tokens and constraining them with masks. 
However, it still needs to optimize motion tokens for each reference video~\cite{li2026fleximmt}. 
Recent feed-forward approaches avoid such repeated optimization. DisMo~\cite{resslerantal2025dismo} learns appearance-agnostic motion representations for open-world transfer. 
SCAIL-2~\cite{yan2026scail2} directly conditions generation on a driving video, and uses an in-context mask to bind driving and target characters across single- and multi-character animation. 
Nevertheless, DisMo does not provide target-region routing for multiple motion sources, while SCAIL-2 obtains multi-character motions jointly from a driving video rather than composing motions from multiple independent reference videos. 
In our work, RART extracts each motion, anchors it with its target region, and inject it into the temporally corresponding latent frames, without per-video optimization.

\section{Methodology}

\subsection{Task Formulation}

Given a target first-frame image $I^{\mathrm{tgt}}$, a text condition $C$, and $N$ object-wise control signals, the $i$-th control signal contains a reference video $V_i^{\mathrm{ref}}$, its per-frame mask $M_i^{\mathrm{ref}}$, and a target mask $M_i^{\mathrm{tgt}}$ on $I^{\mathrm{tgt}}$. 
RART aims to generate a target video $V^{\mathrm{tgt}}$ in which the region specified by $M_i^{\mathrm{tgt}}$ follows the motion of $(V_i^{\mathrm{ref}},M_i^{\mathrm{ref}})$ while preserving the target appearance and background, which can be formulated as:
\begin{equation}
\label{eq:task}
V^{\mathrm{tgt}} = G_{\theta}\!\left(
I^{\mathrm{tgt}},C,
\{(V_i^{\mathrm{ref}},M_i^{\mathrm{ref}},M_i^{\mathrm{tgt}})\}_{i=1}^{N}
\right),
\end{equation}
where $G_\theta(\cdot)$ denotes RART, $(V_i^{\mathrm{ref}},M_i^{\mathrm{ref}})$ specifies which object motion to extract, and $M_i^{\mathrm{tgt}}$ specifies where it should be applied. 
Superscripts $\mathrm{ref}$ and $\mathrm{tgt}$ distinguish reference and target quantities. Uppercase $M$ denotes pixel-space masks, while lowercase $m$ denotes target masks on the latent grid.
Note that the masks can be obtained from segmentation models \cite{cheng2022mask2former,kirillov2023sam,ravi2024sam2,carion2025sam3}.

\subsection{Overview of RART}

As illustrated in Figure~\ref{fig:method}, RART builds upon a video Diffusion Transformer \cite{peebles2023dit,wanteam2025wan}. For $i$-th control signal, RART extracts its reference motion from $V_i^{\mathrm{ref}}$ with $M_i^{\mathrm{ref}}$, anchors it with the target object mask $M_i^{\mathrm{tgt}}$, and finally produces corresponding motion tokens $Z_i$:
\begin{equation}
\label{eq:overview}
\begin{aligned}
F_i &= E_{\mathrm{mot}}(V_i^{\mathrm{ref}},M_i^{\mathrm{ref}}),\\
Z_i &= E_{\mathrm{anchor}}(F_i,M_i^{\mathrm{tgt}}),\\
\end{aligned}
\end{equation}
where $E_{\mathrm{mot}}(\cdot)$ denotes motion extraction and feature projection into the DiT feature space, and $F_i$ is the projected reference-motion feature sequence. 
$E_{\mathrm{anchor}}$ denotes the function that encodes the reference-motion feature and the target object mask into the motion token $Z_i$. 
For all the attained motion tokens of $N$ objects, we concatenate them along token dimension as $Z= \mathrm{Concat}(Z_1,\ldots,Z_N)$.
We denote the DiT hidden dimension by $d$, and all token features below use this dimension. The frozen DiT reads $Z$ through frame-aligned motion cross-attention, while a motion-attention routing loss directly supervises object-wise assignment during training.

The design separates three roles that are often entangled in implicit motion transfer: the masked reference video provides \emph{what} motion should be transferred, the target object mask specifies \emph{where} the motion should act, and the frozen video generator determines \emph{how} the target scene should be rendered. This separation is especially important for multi-object composition. 

\subsection{Mask-Anchored Motion Conditioning}

\paragraph{Motion Tokenization.}
As shown in Figure~\ref{fig:method}, for the $i$-th control signal, the reference mask $M_i^{\mathrm{ref}}$ suppresses unrelated regions in $V_i^{\mathrm{ref}}$ before motion extraction. RART then uses the DisMo encoder \cite{resslerantal2025dismo} as a pretrained motion feature extractor to obtain the temporally ordered sequence $F_i$. Since these features describe reference motion but do not specify where it should act, we first encode the target first-frame mask using mask encoder $E_{\mathrm{mask}}(\cdot)$ as:
\begin{equation}
A_i=E_{\mathrm{mask}}(M_i^{\mathrm{tgt}}).
\end{equation}
The mask encoder consists of convolutional blocks, adaptive pooling and a linear projection to the DiT feature space. Its output $A_i$ is a compact set of spatial tokens that encodes the designated target region. We then inject this spatial anchor into the temporally ordered motion features and group the resulting tokens according to the latent video frames:
\begin{equation}
Z_i=\mathrm{Chunk}\!\left(
\mathrm{MaskCA}((F_i + P_e),A_i)
\right).
\end{equation}
Here, $P_e$ denotes a sinusoidal temporal encoding. $\mathrm{MaskCA}(\cdot)$ denotes a residual cross-attention operation, with the projected motion features as queries, and mask tokens $A_i$ as keys and values. This associates the motion features with the designated target region while preserving their temporal order. $\mathrm{Chunk}(\cdot)$ groups consecutive features according to video latent temporal length, producing $Z_i\in\mathbb{R}^{T\times K\times d}$, where $T$ is the number of latent frames and $K$ is the number of motion tokens per object at each latent frame. The control signals are encoded independently, and then concatenated for frame-aligned motion cross-attention.

\paragraph{Frame-Aligned Motion Cross-Attention.}
Given the video latent $H\in\mathbb{R}^{T\times P\times d}$ in DiT, and the concatenated motion tokens $Z\in\mathbb{R}^{T\times NK\times d}$ of $N$ objects, we conduct the cross-attention between $H$ and $Z$ of each corresponding time slot for motion condition injection. Such frame-aligned motion cross-attention is formulated as: 
\begin{equation}
\label{eq:frame_aligned_mca}
H_f'=\mathrm{CA}(H_f,Z_f), f \in [1, T]
\end{equation}
where $\mathrm{CA}(\cdot)$ is the cross-attention operation with a zero-initialized output projection to preserve the pretrained generator at the beginning of training.
$H_f$ provides the queries, while $Z_f$ provides the keys and values for the $f$-th time slot. 
Each sample--frame pair is processed independently, so the $P$ latent tokens of a frame can access only the corresponding $NK$ motion tokens, for the $K$ tokens from each object. Object-wise tokens are concatenated only within the same frame, allowing multiple motions to be injected jointly without mixing motion tokens across frames; the updated features are then restored to the original layout.

\subsection{Compositional Training}

The preceding modules determine how each motion enters the generator; training must further enforce correct composition across multiple active objects.
For each multi-object training clip, we sample an active object subset $\mathcal{S}$ and concatenate its motion tokens as $Z_{\mathcal{S}}$. Let $m_i$ denote target mask sequence of object $i$ on the latent grid. These sequences provide training-only supervision; no target mask sequence is needed at inference beyond $M_i^{\mathrm{tgt}}$. We also apply frame-consistent spatial augmentation to the reference video--mask pairs during training.

\paragraph{Foreground-Weighted Flow Matching.}
The active target masks define the foreground weights for the flow-matching objective \cite{lipman2023flowmatching}. Let $x$ be the latent encoding of $V^{\mathrm{tgt}}$, $\epsilon$ be Gaussian noise, $\tau$ be a sampled flow time, and $x_\tau=(1-\tau)x+\tau\epsilon$. The target velocity is $v=\epsilon-x$, and the predicted velocity is $\hat{v}_\theta=v_\theta(x_\tau;I^{\mathrm{tgt}},C,Z_{\mathcal{S}})$. For the active object set $\mathcal{S}$, we define
\begin{equation}
\label{eq:flow_objective}
\begin{aligned}
w_{\mathcal{S}}
&= \lambda_b+(1-\lambda_b)
\max_{i\in\mathcal{S}}m_i,\\
\mathcal{L}_{\mathrm{fm}}
&= \mathbb{E}_{x,\tau,\epsilon}
\left[
\frac{\sum w_{\mathcal{S}}\odot
(\hat{v}_\theta-v)^2}
{\sum w_{\mathcal{S}}}
\right].
\end{aligned}
\end{equation}
In Eq.~\eqref{eq:flow_objective}, $w_{\mathcal{S}}$ is broadcast over latent channels, and the sums cover all latent elements. Foreground positions receive unit weight, while background positions receive $\lambda_b$, with $0<\lambda_b<1$. We also apply motion dropout by replacing $Z_{\mathcal{S}}$ with zeros with a small probability, enabling classifier-free-style motion guidance at inference.

\begin{figure}[t]
\centering
\includegraphics[width=1\columnwidth]{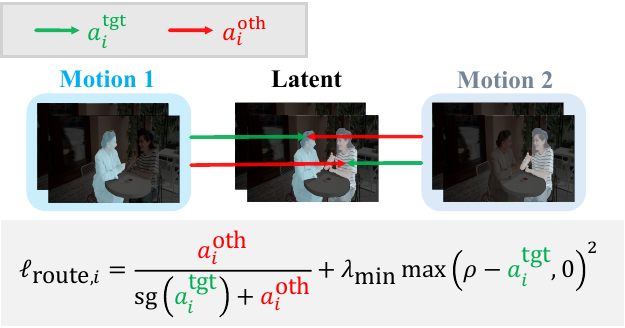}
\caption{Motion-Attention Routing Loss. For each object, RART encourages its target region to attend to the corresponding motion tokens while discouraging other object regions from attending to the same token block.}
\label{fig:route_regularization}
\end{figure}

\paragraph{Motion-Attention Routing Loss.}
Reconstruction alone does not specify which region should read which motion tokens. For multi-object samples, we regularize the motion-attention maps so that each object region routes to its corresponding motion tokens. Figure~\ref{fig:route_regularization} illustrates this operation.
For object $i$, let $a_i^{\mathrm{tgt}}$ denote the attention mass from latent positions within its target-mask sequence $m_i$ to the corresponding motion-token block, and let $a_i^{\mathrm{oth}}$ denote the attention mass from the other active object regions to the same block. Both quantities are obtained directly from the motion-attention maps and averaged over heads, frames, and valid spatial positions. The per-object routing loss is
\begin{equation}
\label{eq:route_loss}
\begin{aligned}
\ell_{\mathrm{route},i}
= \frac{a_i^{\mathrm{oth}}}
{\mathrm{sg}(a_i^{\mathrm{tgt}})+a_i^{\mathrm{oth}}}
+ \lambda_{\mathrm{min}}
\max(\rho-a_i^{\mathrm{tgt}},0)^2.
\end{aligned}
\end{equation}
Here $\mathrm{sg}(\cdot)$ stops gradients through the target-region term in the denominator. The first term penalizes attention routed from other object regions, while the second is activated when $a_i^{\mathrm{tgt}}<\rho$ to prevent all motion-token attention from collapsing. The weight $\lambda_{\mathrm{min}}$ controls this minimum-attention constraint, and $\rho$ is the target attention threshold. Averaging Eq.~\eqref{eq:route_loss} over active objects and motion-attention modules gives $\mathcal{L}_{\mathrm{route}}$. The final objective is
\begin{equation}
\label{eq:total_objective}
\begin{aligned}
\mathcal{L}_{\mathrm{total}}
&= \mathcal{L}_{\mathrm{fm}}
+ \lambda_{\mathrm{route}}\mathcal{L}_{\mathrm{route}},
\end{aligned}
\end{equation}
where $\lambda_{\mathrm{route}}$ balances the routing loss. The routing loss in Eq.~\eqref{eq:total_objective} is applied only to non-dropped multi-object training batches and is obtained from the same forward-pass attention maps, without an additional denoiser evaluation. It therefore directly supervises object-to-motion assignment rather than relying on reconstruction alone.

\begin{table*}[t]
\centering
\begin{tabular}{lccccccccc}
\toprule
& \multicolumn{5}{c}{Automatic Metrics}
& \multicolumn{3}{c}{Human Evaluations (\%)} & \\
\cmidrule(lr){2-6}\cmidrule(lr){7-9}
Method
& TS$\uparrow$
& TC$\uparrow$
& AC$\uparrow$
& FF$\uparrow$ & MF$\uparrow$
& App.$\uparrow$ & Temp.$\uparrow$ & Motion$\uparrow$ & Time$\downarrow$ \\
\midrule
AnyV2V
& 0.261 & 0.879 & 0.742
& 0.571 & 0.338 & 0.0 & 0.0 & 0.0 & 176s/ref + 50s \\
DisMo
& 0.266 & \textbf{0.944} & \textbf{0.913}
& 0.652 & 0.451 & 3.0 & 3.0 & 2.6 & 113s \\
SCAIL-2
& 0.266 & 0.922 & 0.906
& 0.648 & 0.428 & 15.8 & 9.0 & 24.1 & 233s \\
FlexiMMT
& \textbf{0.267} & 0.925 & 0.899
& 0.671 & 0.462 & 14.7 & 27.1 & 21.8 & 2h/ref + 432s \\
\midrule
\textbf{RART}
& 0.265 & 0.928 & 0.903
& \textbf{0.713} & \textbf{0.480} & \textbf{66.5} & \textbf{60.9} & \textbf{51.5} & \textbf{41s} \\
\bottomrule
\end{tabular}
\caption{Quantitative and human evaluations. TS, TC, AC, FF, and MF denote text similarity, temporal consistency, appearance consistency, flow fidelity, and motion fidelity, respectively. Human evaluation columns report five-way selection rates (\%), while Time denotes runtime. For AnyV2V and FlexiMMT, the first time term denotes per-reference preparation.}
\label{tab:main_results}
\end{table*}

\begin{figure*}[t]
\centering
\includegraphics[width=0.95\textwidth]{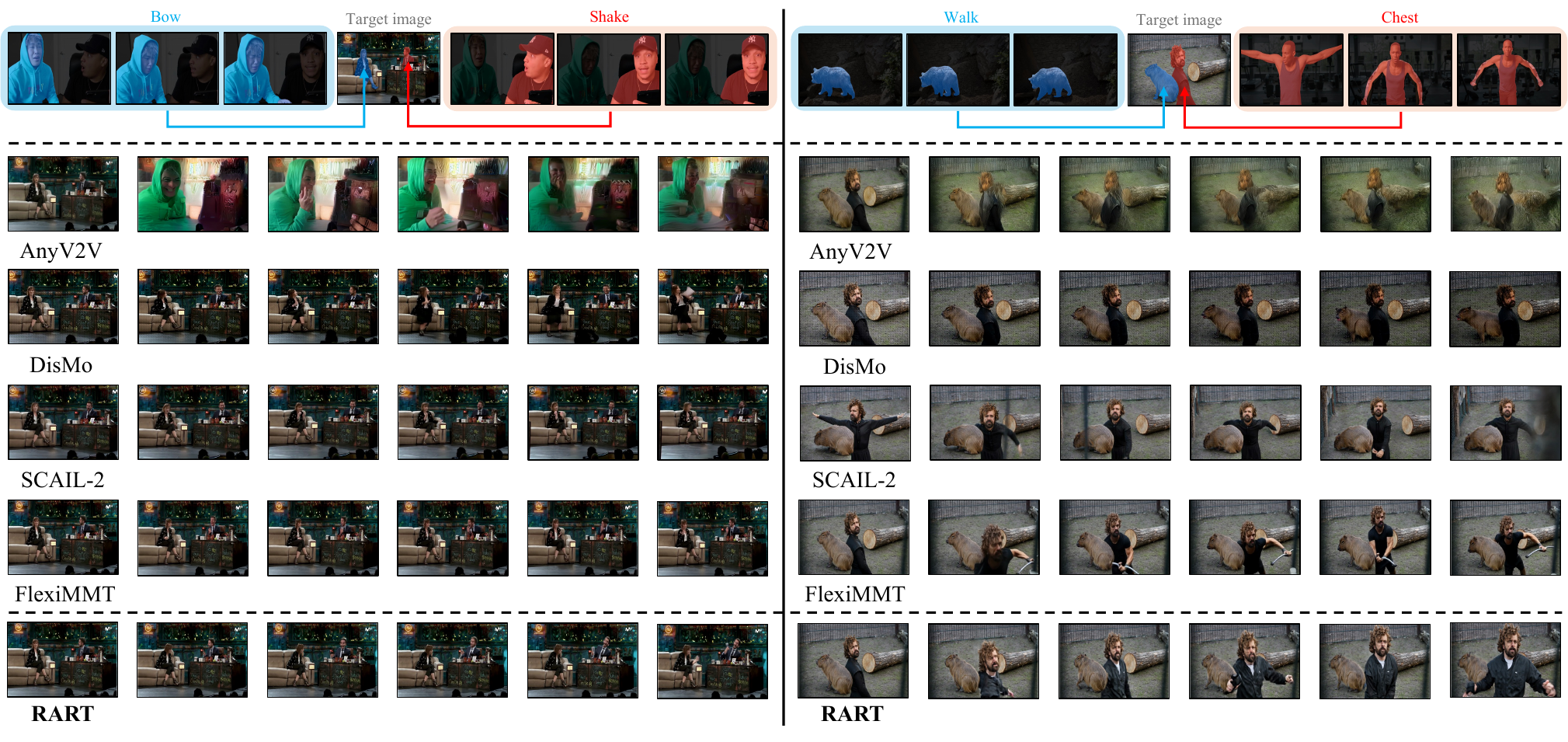}
\caption{Qualitative comparison. The top row shows the reference motions, target image, and motion-to-object assignments; the remaining rows show frames generated by the baselines and RART. RART transfers each reference motion to its designated target object while preserving target appearance.}
\label{fig:qualitative}
\end{figure*}

\section{Experiments}

In this section, we report quantitative and human evaluation results in Table~\ref{tab:main_results}, qualitative comparisons in Figure~\ref{fig:qualitative}, and component analyses in Table~\ref{tab:ablation}, Figure~\ref{fig:ablation}, and Figure~\ref{fig:route_heatmap}.

\subsection{Experimental Setup}

\subsubsection{Datasets.}
We train RART on a mixture of OpenVidHD \cite{nan2025openvid} and HumanVid \cite{wang2024humanvid} videos. After filtering invalid and low-quality samples, the training set contains 286,573 video entries. We extract per-frame object masks with Grounding-SAM2 \cite{liu2024groundingdino,ravi2024sam2}. For evaluation, we extend the FlexiMMT test set \cite{li2026fleximmt} to a multi-object benchmark with 266 test cases and 25 reference videos. Each test sample consists of a target first frame, a text prompt, target first-frame object masks, and multiple reference video-mask pairs. Each target mask specifies where a motion should be applied in the target image, while the corresponding reference-mask sequence marks the object providing that motion in the reference video.

\subsubsection{Baselines.}
We compare RART with four representative implicit methods closely related to motion transfer. AnyV2V \cite{ku2024anyv2v} is a training-free method that injects source-video features into an image-to-video process. DisMo \cite{resslerantal2025dismo} is a feed-forward method that learns appearance-agnostic motion representations for open-world motion transfer. SCAIL-2 \cite{yan2026scail2} is a character-animation method that directly conditions generation on driving videos and uses in-context masks to bind driving and target characters. FlexiMMT \cite{li2026fleximmt} is the closest multi-object multi-motion baseline, which learns object-wise motion tokens under mask constraints but requires reference-specific optimization.

To evaluate all baselines in our setting, we implement the necessary task-specific adaptations while retaining their released backbone weights and core generation mechanisms. For AnyV2V, we independently extract intermediate features from each reference video, fuse them while preserving their mean and variance, and use the fused features for image-to-video inference conditioned on the target image. For DisMo, we independently extract the motion embedding of each reference and fuse the embeddings before its CogVideoX generator. For SCAIL-2, we assemble the independently masked reference clips into a unified multi-character driving sequence and use color-coded masks to preserve object correspondence. As for FlexiMMT, we use an object-specific text description of each designated object during training to distinguish the motion sources in multi-object reference videos. All methods receive the same target image and reference videos. Object assignments are provided directly when supported; otherwise, they are encoded through the task-specific adaptations described above. Apart from these adaptations, all other configurations follow the respective original papers.

\subsubsection{Evaluation metrics.}
We mainly follow the evaluation protocol of FlexiMMT \cite{li2026fleximmt}. Specifically, we report CLIP-based text similarity, temporal consistency, and appearance consistency, together with two object-wise motion metrics: flow fidelity and motion fidelity. Flow fidelity compares mask-restricted optical-flow statistics between the reference object and the generated target object, while motion fidelity compares tracking trajectories inside the corresponding masks. For multi-object samples, we compute object-level scores and average them over all controlled objects.
We additionally conduct human evaluation along three dimensions: target-appearance consistency, temporal consistency, and reference-motion consistency. For each of the 266 test cases, 5 annotators are shown the target first frame, all reference videos, and the outputs of five methods in randomized order. They independently select the best result for each criterion. We aggregate the selections over all test cases and report per-method selection rates, which sum to 100\% for each criterion. We also report runtime to assess efficiency.

\subsubsection{Implementation details.}
RART is built on a frozen Wan2.2 TI2V-5B backbone \cite{wanteam2025wan}. We use the DisMo encoder as the pretrained motion feature extractor \cite{resslerantal2025dismo}, and train only the motion encoding and motion-attention injection modules. The final model is trained for 200K steps with a learning rate of $1\times10^{-5}$ under our mixed-object training configuration. We sample 49 frames from each training video. Training begins at $320\times480$ resolution and is increased to $480\times736$ at 120K steps for high-resolution finetuning. We use 4 motion tokens per latent frame and 2 residual target-mask cross-attention layers. During training, we sample 2 to 6 active objects for multi-object clips, use motion dropout with probability 0.1, caption dropout with probability 0.1, and apply mask perturbation with probability 0.6. As an auxiliary augmentation, each reference video--mask pair undergoes an independent spatial transform shared across frames, while the target condition remains unchanged. We set the background weight to $\lambda_b=0.1$. The motion-attention routing loss is enabled from step 170K with $\lambda_{\mathrm{route}}=0.2$, $\lambda_{\mathrm{min}}=0.05$, and $\rho=0.3$. All experiments are conducted on eight NVIDIA H200 GPUs.

\subsection{Comparisons with State-of-the-Art Methods}

\subsubsection{Quantitative Evaluation.}
Table~\ref{tab:main_results} presents the quantitative results. RART achieves the best flow fidelity and motion fidelity, demonstrating more accurate transfer of reference motions to the designated target objects. It also obtains competitive text similarity, temporal consistency, and appearance consistency, showing that the improvement in motion accuracy does not compromise overall video quality.

The latter three metrics should nevertheless be interpreted together with motion fidelity. Temporal consistency and appearance consistency are computed from frame-to-frame similarities and may therefore favor videos with limited motion, while text prompts rarely describe fine-grained object dynamics and therefore make text similarity an incomplete measure of motion transfer. Consequently, strong scores on these generic video-quality metrics are less informative when the generated motion does not follow the reference.

RART also has the shortest runtime. AnyV2V performs additional per-video noising and caches intermediate features; FlexiMMT optimizes motion tokens for each reference motion and includes an additional mask-extraction stage at inference; and SCAIL-2 uses a larger generative backbone. DisMo remains efficient, but its original formulation does not support compositional multi-object motion transfer. Overall, RART provides the strongest balance of motion-transfer accuracy, visual quality, and inference efficiency.

\subsubsection{Human evaluation.}
The human evaluation results in Table~\ref{tab:main_results} show that RART achieves the highest selection rate across all three criteria, receiving 66.5\% for target-appearance consistency, 60.9\% for temporal consistency, and 51.5\% for reference-motion consistency. These results confirm that RART transfers reference motions more accurately while maintaining coherent dynamics and target appearance.

\subsubsection{Qualitative results.}
Figure~\ref{fig:qualitative} compares the generated videos of different methods. AnyV2V fails to perform multi-motion transfer and produces videos with poor visual quality. DisMo better preserves video quality, but its architecture lacks explicit object-wise spatial assignment and therefore cannot transfer the desired motions precisely. FlexiMMT can transfer motions extracted from single-object reference videos, but struggles to isolate and transfer the motion of a designated region when the reference video contains multiple objects. SCAIL-2 does not natively support extracting motions from multiple reference videos; although our adaptation enables this input setting, its generated videos often drift from the target first frame and exhibit weak temporal consistency. In contrast, RART consistently extracts the specified motions, assigns them to the intended target objects, and preserves both target appearance and temporal coherence.

\begin{figure}[t]
\centering
\includegraphics[width=1\columnwidth]{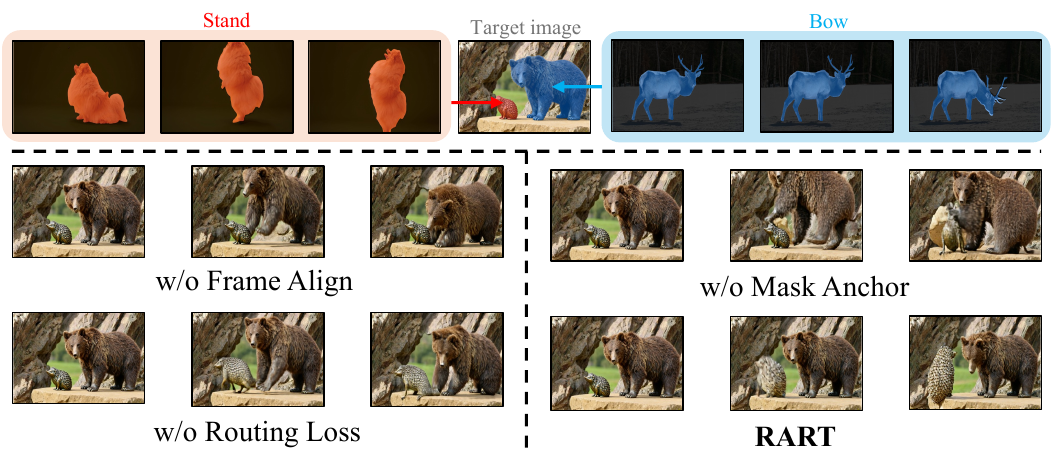}
\caption{Qualitative ablation. The top row shows reference motions and their assignments to target regions; the bottom panels compare RART with variants without frame alignment, target-mask anchoring, or routing loss.}
\label{fig:ablation}
\end{figure}

\begin{figure}[t]
\centering
\includegraphics[width=1\columnwidth]{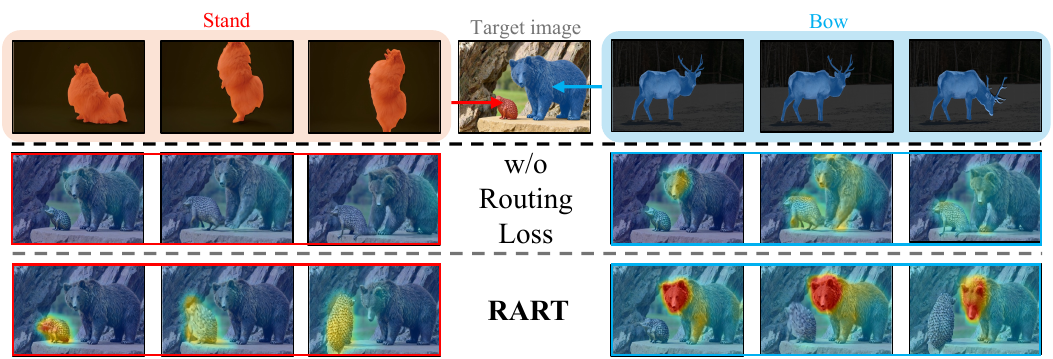}
\caption{Motion-attention heat maps. Columns correspond to the two reference motions, while rows compare the model without the routing loss and RART. Warmer colors indicate stronger attention to each motion-token block.}
\label{fig:route_heatmap}
\end{figure}

\begin{table}[t]
\centering
\begin{tabular}{lcc}
\toprule
Variant & FF$\uparrow$ & MF$\uparrow$ \\
\midrule
w/o Frame Alignment & 0.678 & 0.400 \\
w/o Mask Anchor & 0.690 & 0.413 \\
w/o Routing Loss & \textbf{0.727} & 0.448 \\
\midrule
\textbf{RART} & 0.713 & \textbf{0.480} \\
\bottomrule
\end{tabular}
\caption{Quantitative ablations. We replace frame-aligned motion attention with global attention, use all-ones target masks, or remove the motion-attention routing loss.}
\label{tab:ablation}
\end{table}

\subsection{Ablation Study}

\subsubsection{w/o Frame Alignment.}
We train this variant by replacing frame-aligned motion cross-attention with global motion cross-attention while keeping all other settings unchanged. As shown in Figure~\ref{fig:ablation}, neither the bear nor the hedgehog follows the temporal motion pattern of its reference video, whereas the full model with frame-aligned motion cross-attention generates correctly aligned motions. Consistently, Table~\ref{tab:ablation} shows that this variant obtains lower FF and MF scores than the full model. These results demonstrate that frame-aligned motion cross-attention is essential for establishing temporal correspondence.

\subsubsection{w/o Target-Mask Anchoring.}
To evaluate the effect of target-mask anchoring, we replace every target mask $M_i^{\mathrm{tgt}}$ with an all-ones mask at inference. Consequently, all objects receive the same mask condition and no longer have object-specific target-region information. As shown in Figure~\ref{fig:ablation}, the bear incorrectly follows the motion intended for the hedgehog when the spatial anchors are removed. Table~\ref{tab:ablation} also reports lower FF and MF scores for this variant than for the full model. Together, these results demonstrate that target-mask anchoring is crucial for assigning each reference motion to its intended target region.

\subsubsection{w/o Motion-Attention Routing Loss.}
We continue training from step 170K without the motion-attention routing loss, while keeping all other settings unchanged. Figure~\ref{fig:ablation} shows that, although the hedgehog partially follows its reference motion, the bear incorrectly performs the motion intended for the hedgehog, revealing cross-object motion leakage.

Table~\ref{tab:ablation} further quantifies this effect. The routing loss raises MF from 0.448 to 0.480, a 7.1\% relative improvement, with only a small FF change from 0.727 to 0.713. FF captures local flow magnitude and direction, whereas MF measures trajectory-level correspondence, making them complementary indicators of motion transfer. The full model therefore retains strong local flow fidelity while substantially improving trajectory-level correspondence.

To examine the routing behavior directly, we further visualize the attention maps in Figure~\ref{fig:route_heatmap}. For each object, we aggregate the weights assigned to its motion-token block at every latent spatial position and upsample the resulting heat map to the original video resolution. With the routing loss, attention to each object's motion tokens is concentrated in its designated target region. Without the loss, the attention maps of different motion sources become substantially mixed. Together with the qualitative and quantitative results, these heat maps show that the routing loss improves object-wise motion assignment and suppresses cross-object leakage.

\section{Conclusions}

We presented RART, a unified framework for compositional multi-object motion transfer. RART represents each reference motion with object-wise tokens anchored by the corresponding target mask, and injects temporally matched tokens into a frozen video Diffusion Transformer through frame-aligned motion cross-attention. The motion-attention routing loss further strengthens the correspondence between target regions and motion sources, reducing cross-object motion leakage. RART jointly composes multiple reference motions in one denoising pass, without separate optimization for each reference video. Quantitative, qualitative, and human evaluations show that RART delivers accurate motion transfer, strong visual quality, and efficient inference; ablations and attention visualizations support our key design choices. Future work will explore broader applications of RART in long-form generation and interactive multi-object video creation.

\bibliography{rart_refs,rart_supp_refs}

\clearpage
\appendix

This document provides additional implementation details, dataset and evaluation protocols, baseline adaptations, additional qualitative results, and liminations of RART. The main paper contains all information essential to the central claims; this document supplies supporting material and reproducibility details.

\section{Training and Inference Details}

\subsection{Training Details}

Table~\ref{tab:supp_training_config} summarizes the training configuration used for the reported checkpoint. RART uses a frozen Wan2.2 TI2V-5B backbone \cite{wanteam2025wan} and the pretrained DisMo motion extractor \cite{resslerantal2025dismo}.

\begin{table}[t]
\centering
\begin{tabular}{@{}p{0.6\columnwidth}p{0.35\columnwidth}@{}}
\toprule
Configuration & Value \\
\midrule
Video backbone & Wan2.2 TI2V-5B \\
Motion extractor & DisMo Encoder \\
Training steps & 200K \\
Learning rate & $1\times10^{-5}$ \\
Hardware & 8 $\times$ NVIDIA H200 GPUs \\
Batch size per GPU & 1 \\
Optimizer & AdamW \\
Weight decay & $1\times10^{-4}$ \\
Warmup steps & 4000 \\
Training frames & 49 \\
Initial resolution & $320\times480$ \\
Final resolution & $480\times736$ \\
Resolution switch & 120K steps \\
Motion tokens per frame & 4 \\
Target-mask attention layers & 2 \\
Active objects per clip & 1--6 \\
Motion dropout & 0.1 \\
Caption dropout & 0.1 \\
Mask perturbation probability & 0.6 \\
$\lambda_b$ & 0.1 \\
$\lambda_{\mathrm{route}}$ & 0.2 \\
$\lambda_{\mathrm{min}}$ & 0.05 \\
$\rho$ & 0.3 \\
Routing loss enabled at & 170K steps \\
\bottomrule
\end{tabular}
\caption{Training configuration of the reported RART model.}
\label{tab:supp_training_config}
\end{table}

\subsection{Inference Details}

We use the checkpoint at 200K training steps for all reported results. Each sample is generated at a resolution of $480\times736$ with 49 frames. We use a 50-step Flow-UniPC sampler with a flow shift of 5.0. The text and motion guidance scales are set to 3.0 and 5.0, respectively, and the random seed is fixed to 42 for all test samples.

At each sampling step, let $\hat{v}_{u}$ denote the text-unconditional prediction, $\hat{v}_{c}$ the text-conditioned prediction without motion, and $\hat{v}_{c,m}$ the prediction with both text and motion conditions. We combine them as
\[
\hat{v}=\hat{v}_{u}
+3.0(\hat{v}_{c}-\hat{v}_{u})
+5.0(\hat{v}_{c,m}-\hat{v}_{c}).
\]

Reference videos and their masks are sampled to 49 frames and resized to the generation resolution. Before motion extraction, pixels outside each reference mask are set to zero. Target masks are resized to the latent grid and passed to the mask-encoder module. All reference motions are extracted independently and injected jointly without reference-specific optimization.

\section{Dataset and Evaluation Details}

\subsection{Training Set Construction}

We construct our data pool from OpenVidHD \cite{nan2025openvid} and HumanVid \cite{wang2024humanvid}. For both datasets, we first retain videos containing at least 49 frames and labeled with static camera motion. We then use Qwen3-VL-8B-Instruct \cite{bai2025qwen3vl} to re-caption each video from its original caption and uniformly sampled frames. The model returns a refined caption together with a list of main subjects and their corresponding actions. Subjects labeled as static are removed because they do not provide useful motion supervision; videos without any remaining moving subject are excluded from the subsequent segmentation stage.

We next obtain an object mask sequence for every retained subject using Grounding DINO and SAM2 \cite{liu2024groundingdino,ravi2024sam2}. Given the subject description, Grounding DINO detects the subject in the first frame, and its highest-confidence bounding box initializes the SAM2 image mask. SAM2 then propagates this mask throughout the video, producing a temporally aligned mask sequence for each object. We use box and text thresholds of 0.3 and 0.25, respectively, and discard the video if any annotated subject cannot be detected or segmented. After preprocessing and filtering, we obtain 287,540 valid videos with object-level motion descriptions and per-frame masks. We use 286,573 videos for training and reserve the remaining 967 videos as a held-out pool for constructing the test set.

\subsection{Test Set Construction}

We construct our test set by extending the FlexiMMT test set \cite{li2026fleximmt}. From the 967 held-out videos, we select 5 reference video pair and 171 target image pairs and combine them with the original FlexiMMT test data. The resulting test set contains 266 test cases and 25 reference videos in total. None of the newly selected reference videos or target images is used for training RART. Each test case contains a target image, a text prompt, target first-frame masks, and reference video--mask pairs, where the per-frame reference masks identify the objects whose motions should be transferred.

\subsection{Automatic Metrics}

We follow the automatic evaluation protocol implemented in FlexiMMT \cite{li2026fleximmt} and report five metrics, for which higher values are better. Before computing the object-wise motion metrics, we initialize SAM2 \cite{ravi2024sam2} with each target first-frame mask and propagate it through the generated video to obtain per-frame object masks. For multi-object samples, FF and MF are first averaged over all controlled objects. All five metrics are averaged over the test set.

\paragraph{CLIP-Based Metrics.} We use CLIP ViT-B/32 \cite{radford2021clip} to compute Text Similarity (TS), Temporal Consistency (TC), and Appearance Consistency (AC). TS is the mean cosine similarity between the caption embedding and the CLIP image embeddings of all generated frames. TC measures inter-frame consistency using cosine similarities between CLIP image embeddings. AC averages the cosine similarity between the first-frame embedding and each subsequent-frame embedding to measure appearance preservation throughout the video.

\paragraph{Flow Fidelity.} We use RAFT \cite{teed2020raft} to estimate frame-to-frame optical flow in both the reference and generated videos. For each controlled object, FF compares the normalized magnitude and direction distributions of the mask-restricted reference and generated flows using the Bhattacharyya coefficient. The resulting scores are first averaged across corresponding frame pairs and then across controlled objects.

\paragraph{Motion Fidelity.} We use CoTracker \cite{karaev2024cotracker} to extract point trajectories from the reference and generated videos, initialized within the first-frame bounding region of each object. For each tracked point, we compute its displacement from one frame to the next and normalize each displacement vector to unit length. We then construct a trajectory-similarity matrix by averaging directional cosine similarity over frames for every generated-reference trajectory pair. For each generated trajectory, we select the highest-scoring reference trajectory and average these best-match scores over generated trajectories and controlled objects to obtain MF.

\subsection{Human Evaluation Protocol}

We conduct a human evaluation comparing five models across 266 test cases. Each test case is independently evaluated by five annotators. Annotators are shown the target first frame, all reference videos, and five generated outputs. They are not informed of the method identities, and the output order is randomized for each test case. For each criterion, each annotator selects exactly one best result; ties are not allowed.

We evaluate three criteria. Target-appearance consistency measures how well the generated video preserves the objects and visual content of the target first frame. Temporal consistency evaluates whether the generated video remains coherent over time without noticeable flicker or abrupt deformation. Reference-motion consistency assesses how accurately the controlled objects follow their assigned reference motions. For each criterion, we report each model's share of selections, which sums to 100\% across the five models.

\section{Baseline Adaptations}

Existing motion-transfer methods do not natively support RART's input format, which consists of multiple independent reference video--mask pairs and their target-object assignments. We therefore adapt only the conditioning interface needed to represent our task, while retaining each method's released backbone weights, core conditioning mechanism, and sampling procedure.

\subsection{AnyV2V}

The original AnyV2V pipeline \cite{ku2024anyv2v} performs DDIM inversion for one source video and injects source convolutional features together with spatial- and temporal-attention queries and keys during I2VGen-XL denoising. For a sample with multiple reference videos, we run the official inversion and feature-extraction procedure independently for every reference. Features from the corresponding injection sites are then combined using moment-preserving fusion, which maintains the mean and variance of the reference features. We use the target image as the edited first frame and inject the fused features in the image-to-video sampling pass. Because AnyV2V has no target-region routing interface, this fused condition is applied globally. The released I2VGen-XL backbone, injection schedule, and sampling configuration are otherwise unchanged.

\subsection{DisMo}

The original DisMo generation pipeline \cite{resslerantal2025dismo} encodes one motion video into a temporally ordered motion-embedding sequence and supplies it to a motion-conditioned CogVideoX-5B-I2V generator. We extend its sampling interface to accept precomputed motion embeddings. Each reference video is first processed independently by the released DisMo encoder, after which the resulting sequences are averaged element-wise and passed through this interface together with the target image and prompt. This permits several references to condition one generation pass, but the averaged embedding remains a global condition without object-specific target-mask routing. All released motion-extractor and CogVideoX adapter weights, as well as the original sampling procedure, are retained. This baseline differs from RART, which uses the pretrained DisMo encoder only as a frozen feature extractor and learns separate mask-anchored encoding and injection modules.

\subsection{SCAIL-2}

SCAIL-2 \cite{yan2026scail2} takes a target image with a color-coded target mask and a single reference video with a frame-wise color-coded mask sequence. Matching colors establish object correspondence between the two inputs. To accommodate multiple independent reference videos, we use each reference-mask sequence to isolate the corresponding object from its reference video. Each extracted object is resized and positioned according to its corresponding target mask while preserving its aspect ratio. The transformed objects are then composited frame by frame into a single multi-character reference video. Each object is assigned a distinct color from SCAIL-2's supported mask palette, producing a color-coded driving mask that preserves object correspondence. This adaptation changes only the construction of SCAIL-2 inputs; the released generator and inference procedure remain unchanged.

\subsection{FlexiMMT}

FlexiMMT \cite{li2026fleximmt} uses the training caption to obtain a mask for the described object while learning its motion. To adapt it to a multi-object reference video, we provide a separate caption describing each source object. This allows its original mask-extraction pipeline to identify the designated object and extract its motion from the shared reference video. Apart from this caption-based object specification, we follow the official training and inference procedures without further modification.

\section{Additional Qualitative Results}

\begin{figure*}[t]
\centering
\includegraphics[width=\textwidth]{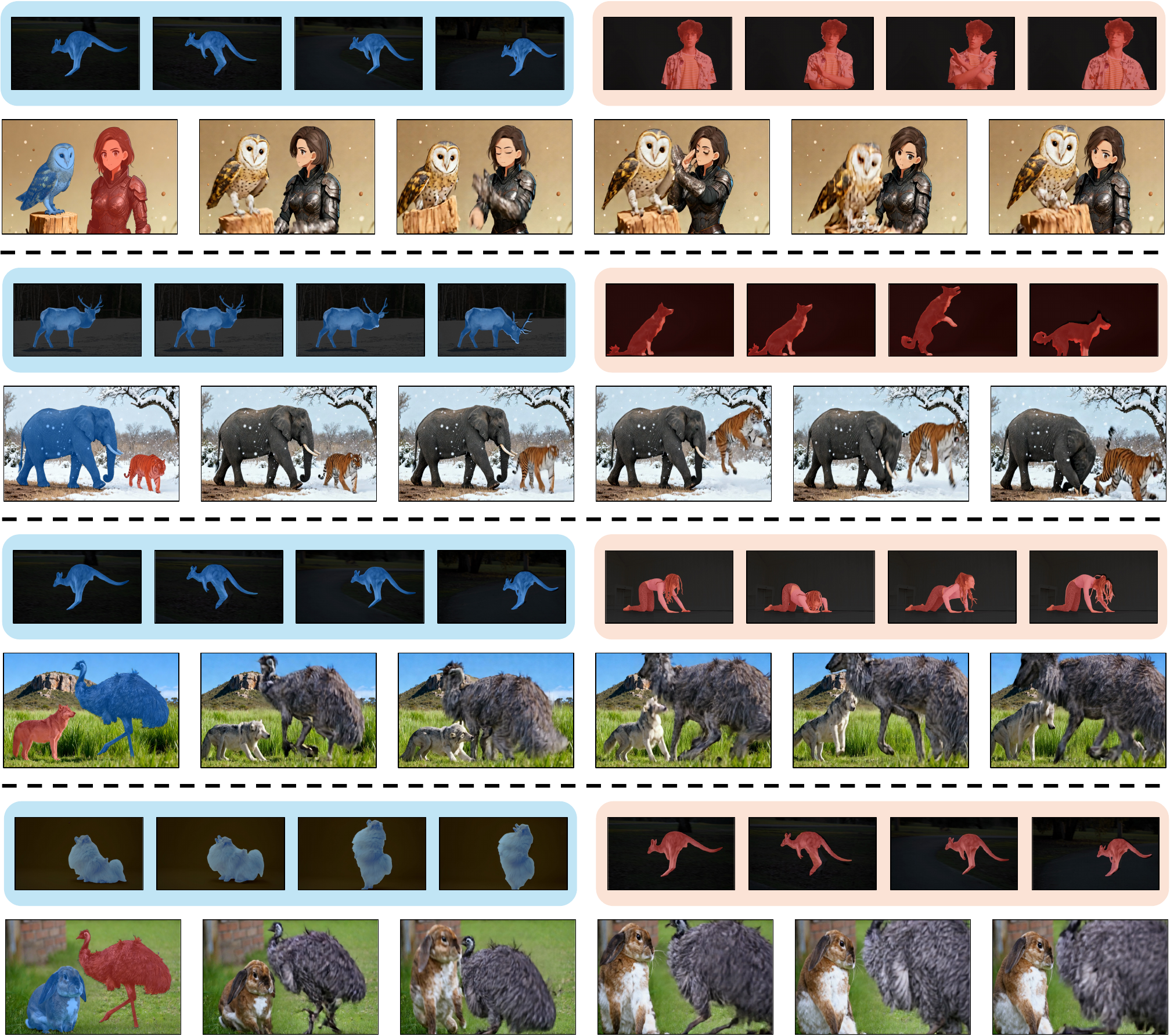}
\caption{Additional qualitative results for compositional motion transfer. Each example shows color-coded reference sequences at the top, followed by the target-object assignments and generated frames at the bottom. Blue and red indicate the correspondence between each reference motion and its target object.}
\label{fig:supp_qualitative_1}
\end{figure*}

\begin{figure*}[t]
\centering
\includegraphics[width=\textwidth]{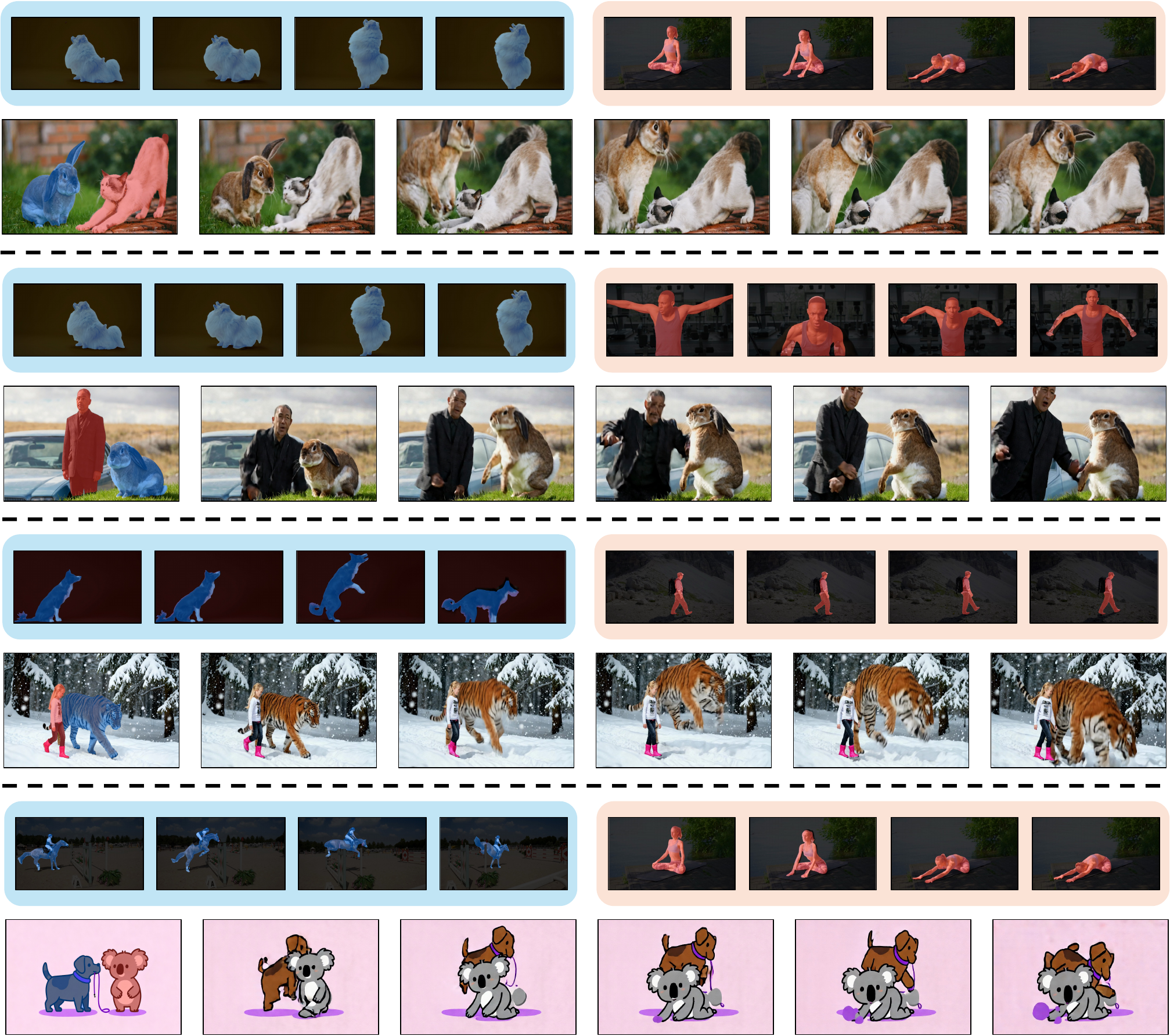}
\caption{Further qualitative results for compositional motion transfer. RART jointly transfers independently specified reference motions to their color-matched target objects while preserving the appearance and scene of the target image.}
\label{fig:supp_qualitative_2}
\end{figure*}

Figures~\ref{fig:supp_qualitative_1} and~\ref{fig:supp_qualitative_2} present additional results for compositional motion transfer. In each example, the blue and red reference sequences provide two independent motions, while the same colors in the target image indicate their assigned target objects. The subsequent frames show the video generated by RART.

\section{Limitations and Future Works}

RART's output quality remains bounded by the capabilities of its frozen video generator and pretrained motion encoder. Because the generator is not updated during training, the added motion-control modules cannot fully compensate for failures in its visual prior. Challenging motions or uncommon target geometries can therefore produce unstable object shapes or visible artifacts. Meanwhile, the motion representation may omit subtle or highly articulated dynamics, reducing motion fidelity in difficult cases.

Future work will explore adapting RART to stronger video generators and more expressive motion encoders. Improving these underlying components should further enhance visual quality and fine-grained motion preservation while retaining RART's object-wise spatial and temporal control.


\end{document}